\documentclass[journal]{IEEEtran}

\usepackage{cite}
\usepackage{amsmath,amssymb,amsfonts}
\usepackage{array}
\usepackage{booktabs}
\usepackage{xcolor}
\usepackage{colortbl}
\usepackage{pifont}
\usepackage{tikz}
\usepackage{tabularx}
\usepackage{dblfloatfix} 
\usepackage{balance}
\usepackage{amsthm}
\usepackage{chngcntr}
\usepackage{graphicx}
\usepackage{textcomp}
\usepackage{url}
\usepackage[hidelinks]{hyperref}

\newtheorem{proposition}{Proposition}
\newenvironment{pf}{\begin{proof}}{\end{proof}}

\def\mlxdraft{0}   
\newcommand\mlxchk[1]{\ifnum\mlxdraft=1\relax\textcolor{red}{#1}\else#1\fi}

\newcolumntype{P}[1]{>{\raggedright\arraybackslash}p{#1}}
\definecolor{posblue}{RGB}{226,238,250}
\newsavebox\mlxpartbox
\AtBeginDocument{\sbox\mlxpartbox{\tikz[baseline=-0.6ex]{\draw[line width=0.45pt,orange!85!black] (0,0) circle (0.72ex);
  \fill[orange!85!black] (0,0.72ex) arc[start angle=90,end angle=270,radius=0.72ex] -- cycle;}}}

\begin{document}



\newcommand{\preprintnote}{Preprint \textemdash\ Submitted for Review}

\title{MambaLIFT: Reference-Based Mechanism-Aware Attribution for VMamba}

\author{Jintu Kakati and Niladri B. Puhan%
\thanks{This manuscript has been submitted for peer review and has not yet been peer reviewed.}%
\thanks{Jintu Kakati and Niladri B. Puhan are with the School of Electrical and Computer Sciences, Indian Institute of Technology Bhubaneswar, Bhubaneswar 752050, India.}%
\thanks{Corresponding author: Niladri B. Puhan (e-mail: nbpuhan@iitbbs.ac.in).}%
}

\IEEEaftertitletext{
\begin{center}
\footnotesize\textit{\preprintnote}
\end{center}
}

\maketitle

\begin{abstract}
Attributing predictions of visual state-space models remains challenging because input-dependent selective dynamics couple feature selection, recurrent state propagation, and shared computational pathways. We introduce MambaLIFT, a reference-based, multiplier-first attribution framework for VMamba that propagates contributions through its computational graph using interaction-aware rules for selective-scan injection, recurrent memory, LayerNorm, and shared pathways. MambaLIFT decomposes the reference-relative class-logit difference, \(f_c(x)-f_c(x^0)\), by spatial position, VSS block, scan direction, and computational mechanism. Experiments on ImageNet-1K evaluate VMamba-T and VMamba-B, with hyperparameters selected on a calibration set and faithfulness assessed on a disjoint held-out evaluation set. MambaLIFT achieves numerical conservation to floating-point precision across audited attribution interfaces in both configurations. Under mean-colour pixel flipping, it achieves significantly higher mean faithfulness scores than MambaLRP on both configurations and the highest scores among the evaluated methods, including Integrated Gradients with matched references. Under blurred replacement, MambaLIFT is non-inferior to MambaLRP on VMamba-T and equivalent on VMamba-B, indicating that its advantage depends on the replacement protocol. Ablations on VMamba-T show that interaction allocation, temporal propagation, and pathway composition affect attribution maps. Mechanism-level analyses further show that exact relevance conservation does not imply additive effects under finite interventions: the mechanism decomposition accounts for the reference-relative logit difference but does not estimate intervention effects or establish causal validity.
\end{abstract}


\section{Introduction}\label{sec:intro}

Visual state-space models such as VMamba~\cite{liu2024vmamba} extend Mamba's selective state-space dynamics~\cite{gu2024mamba} to image recognition. VMamba's SS2D module unfolds feature maps into four scan sequences, processes each with an input-dependent selective scan and merges the outputs. The model has demonstrated strong visual recognition performance, and its linear complexity in sequence length and ability to capture long-range dependencies make it suitable for studying relevance propagation. However, explaining its predictions is challenging: the selective scan combines input-dependent parameters with recurrent state updates, and conventional attribution methods do not explicitly decompose its selection, recurrent-state and shared-pathway contributions.

DeepLIFT~\cite{shrikumar2017deeplift} attributes the difference $f_c(x)-f_c(x^0)$ in the logit $f_c$ for class $c$ between an input $x$ and a reference $x^0$. Applying this principle to VMamba requires accounting for the interactions in its selective recurrence
\begin{equation}
h_t = \bar{A}_t h_{t-1} + \tau_t B_t u_t,
\qquad
\bar{A}_t = \exp(\tau_t A),
\label{eq:recurrence}
\end{equation}
where $h_t$ is the recurrent state at scan step $t$, $\bar{A}_t$ the discretized state-transition matrix, $A$ the state-space parameter matrix, $\tau_t$ the input-dependent step size, $B_t$ the input matrix and $u_t$ the input feature; the term $\tau_t B_t u_t$ injects the input into the state. Their contributions are coupled through the state updates and integrated across scan directions and network blocks, so an attribution method must account for these interactions and shared pathways while separating the contributions of state transitions, input injection and other computational paths.

We propose MambaLIFT, a reference-based, multiplier-first attribution method for VMamba. It adopts DeepLIFT's reference-relative Linear and Rescale principles where appropriate, while introducing symmetric interaction-aware propagation rules for the multiplicative operations specific to selective state-space computation, including pairwise products and the three-factor selective-scan injection term $\tau_t B_t u_t$. Multipliers propagated backward through the computational graph yield input relevance that sums to the logit difference, while contributions from recurrent state propagation, input injection, and scan directions remain separate until their pathways merge.

Existing methods offer complementary approaches to explaining selective state-space models: relevance propagation~\cite{rezaei2024mambalrp}, token-level decomposition~\cite{pitorro2025latim}, implicit-attention analysis~\cite{ali2025hidden} and state-influence analysis~\cite{mabrok2026controllability}. To our knowledge, existing methods do not directly decompose the reference-relative class-logit difference into contributions from VMamba's selection, recurrent state propagation, and scan pathways. MambaLIFT is designed to provide this decomposition. Unlike MambaLRP, which attributes the class logit, MambaLIFT explains the reference-relative logit difference, enabling mechanism-level analysis and conservation audits across VMamba.  It provides a conservation-verified, mechanism-resolved decomposition of VMamba predictions across scan, state-memory, and directional pathways.

The main contributions of this paper are as follows:
\begin{itemize}
    \item MambaLIFT propagates reference-relative multipliers through VMamba's selective-scan operations, recurrent states, LayerNorm, and shared pathways, using established reference-based allocation rules; the resulting input relevance sums to $\Delta f_c$.

    \item It reports relevance by input position, VSS block, scan direction, and computational mechanism, including the content and selection pathways of the selective scan and the part carried by its state memory.

    \item We evaluate MambaLIFT on ImageNet-1K held-out evaluation images with VMamba-T and VMamba-B, examining numerical conservation, gradient-limit behaviour, negative controls, attribution faithfulness, propagation-rule ablations, reference sensitivity and mechanism-level accounting through interventions.
     
\end{itemize}

Section~\ref{sec:related} reviews related work; Section~\ref{sec:method} presents MambaLIFT; Section~\ref{sec:experiments} describes the experimental setup and evaluates conservation and attribution faithfulness; Section~\ref{sec:exp-ablation} reports reference sensitivity, propagation-rule ablations, mechanism-level contributions and computational cost; 
and Section~\ref{sec:conclusion} concludes with directions for future research.
\section{Related work}\label{sec:related}

\subsection{Attribution for vision models}
Gradient-based methods such as Gradient$\times$Input (GI)~\cite{shrikumar2017deeplift}, SmoothGrad~\cite{smilkov2017smoothgrad} and Integrated Gradients (IG)~\cite{sundararajan2017ig} derive attributions from input gradients, smoothed gradients or path-integrated gradients, respectively. Class activation mapping methods instead weight intermediate feature maps; Opti-CAM, for example, optimizes these weights for each image~\cite{zhang2024opticam}. Propagation-based methods, including Layer-wise Relevance Propagation (LRP)~\cite{bach2015lrp,montavon2019lrp}, redistribute output relevance through the network with rules such as LRP-$\gamma$; relevance propagation has also been adapted to attention and LayerNorm in Transformers~\cite{ali2022xai} and guided by attributes learned from intermediate layers~\cite{han2025arp}.

DeepLIFT~\cite{shrikumar2017deeplift} attributes the difference between an input and a reference by propagating activation differences through the network; DeepSHAP~\cite{lundberg2017shap} combines reference-based attribution with Shapley-value approximations, and Expected Gradients~\cite{erion2021expected} averages integrated-gradient attributions over a reference distribution. MambaLIFT builds on DeepLIFT's reference-based principle and adopts a generalized LRP-$\gamma$ rule for convolutions in a reference-relative form that conserves relevance across each convolution.

\subsection{Interaction-aware attribution}
Multiplicative operations require an explicit allocation of interaction contributions among their factors. For a product whose factors change from their reference to their actual values, the symmetric two-factor rule used here coincides with the Shapley-value allocation~\cite{lundberg2017shap}. LRP studies of recurrent networks have also compared propagation rules for the multiplicative gates of LSTMs~\cite{arras2019rnn}. For three-factor products, sequential binary decompositions can yield different allocations depending on how the factors are grouped. Higher-order interaction indices such as Shapley--Taylor~\cite{sundararajan2020shapleytaylor} quantify interactions among input features, whereas MambaLIFT decomposes the reference-relative contribution of the three factors of the selective-scan injection term within VMamba's computational graph.

\subsection{Evaluating attribution methods}

Attribution faithfulness is commonly evaluated by pixel flipping~\cite{samek2017evaluating}, whose outcomes depend on the feature-replacement strategy and the ranking procedure~\cite{bluecher2024decoupling,hooker2019benchmark}. Different quantitative metrics can also grade the same explanations inconsistently~\cite{vandersmissen2024coherency}, and procedures for debiasing attribution evaluation have been proposed~\cite{duan2026rain}. Reference-based attributions also require sensitivity analysis because they can vary with the selected reference~\cite{sturmfels2020baselines,kindermans2019unreliability}. Accordingly, we evaluate MambaLIFT with numerical completeness checks, attribution-map agreement, reference sensitivity and pixel flipping.

\subsection{Explaining state-space models}

Mamba~\cite{gu2024mamba} introduced selective state-space modeling; Vision Mamba (Vim)~\cite{zhu2024vim} extended it to visual sequences, and VMamba~\cite{liu2024vmamba} introduced four-directional SS2D scanning for spatial dependencies. In SS2D, \emph{CrossScan} rearranges the 2-D feature map into sequences along multiple spatial directions, while \emph{CrossMerge} maps the direction-specific outputs back to the spatial feature-map layout. Visual state-space backbones have since been applied to tasks such as medical image segmentation~\cite{chen2025ummamba}. Interpretability methods include MambaLRP~\cite{rezaei2024mambalrp}, which performs conservative relevance propagation for Mamba and Vim while treating selection parameters and normalization denominators as constants; LaTIM~\cite{pitorro2025latim}, which characterizes latent token-to-token interactions; hidden-attention analysis~\cite{ali2025hidden}, which represents selective scanning as an implicit attention matrix; and controllability analysis~\cite{mabrok2026controllability}, which uses Jacobian- and Gramian-based measures to characterize input influence on internal state dynamics.

MambaLIFT addresses a different objective: attributing reference-relative changes in the class logit through VMamba's computational graph. It differs from MambaLRP in three respects. First, MambaLRP distributes \(f_c(x)\), whereas MambaLIFT distributes \(\Delta f_c=f_c(x)-f_c(x^0)\), so additive constants such as biases and LayerNorm shifts cancel (Section~\ref{sec:correctness}). Second, MambaLRP treats \(\tau_t\), \(B_t\), and \(C_t\) as constants, whereas MambaLIFT allocates their input-dependent interactions with \(u_t\) and the state, yielding separate shares for selection, recurrent content, skip, and memory pathways. Third, MambaLIFT retains these shares by scan direction and block and audits conservation at each interface. Table~\ref{tab:method_comparison} summarizes the resulting positioning.

\begin{table*}[t]
\centering
\caption{Positioning of MambaLIFT relative to recent interpretability approaches for state-space models.}
\label{tab:method_comparison}
\small
\setlength{\tabcolsep}{2pt}
\renewcommand{\arraystretch}{1.18}
\begin{tabular}{@{}P{3.0cm}P{4.2cm}P{4.1cm}P{5.1cm}@{}}
\toprule
\textbf{Method} &
\textbf{Model / setting} &
\textbf{What is explained} &
\textbf{Output / analytical resolution} \\
\midrule

MambaLRP \cite{rezaei2024mambalrp}
&
Mamba, Vision Mamba
&
Class logit $f_c(x)$ through layer-wise relevance propagation
&
Relevance per input token or pixel; selective-state-space propagation
\\
\addlinespace

LaTIM \cite{pitorro2025latim}
&
Mamba
&
Token-to-token interactions within Mamba representations
&
Token interaction maps
\\
\addlinespace

Hidden attention \cite{ali2025hidden}
&
Mamba, Vision Mamba
&
Selective scan viewed as implicit attention
&
Attention-like influence maps
\\
\addlinespace

Controllability analysis \cite{mabrok2026controllability}
&
Vision State-Space Models
&
Influence of input positions on internal state dynamics
&
Jacobian/Gramian-based influence scores at patch/state level
\\

MambaLIFT
&
VMamba-T/B
&
Reference-relative class-logit difference
$f_c(x)-f_c(x^0)$
&
Relevance per spatial position, block, scan direction, and computational mechanism through the selective SS2D graph
\\

\bottomrule
\end{tabular}
\end{table*}
\section{Proposed Methodology}\label{sec:method}

\begin{figure*}[!t]
\centering
\includegraphics[width=\textwidth]{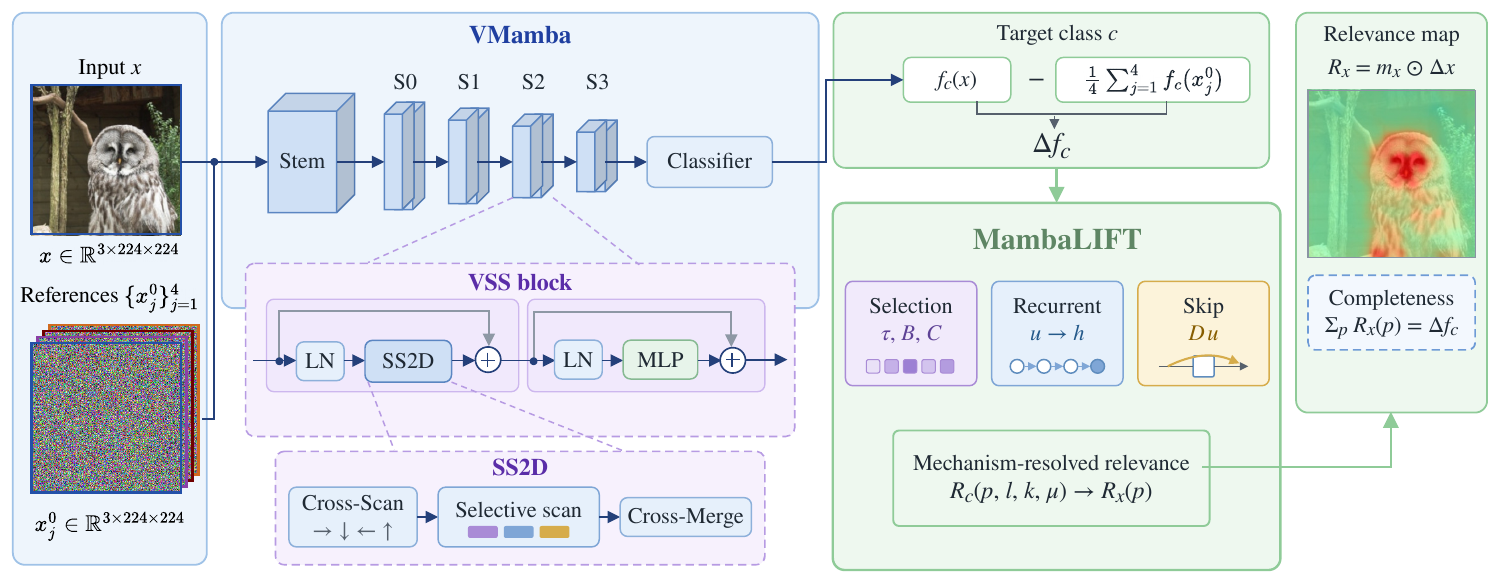}
\caption{Overview of MambaLIFT. The input and reference images pass through the same VMamba, and multipliers are propagated backward through the selective SS2D graph. Relevance is resolved by spatial position, VSS block, scan direction, and computational mechanism.}
\label{fig:mambalift_framework_1}
\end{figure*}

Fig.~\ref{fig:mambalift_framework_1} summarizes MambaLIFT.  The input and reference images are processed by the VMamba to obtain the reference-relative class-logit difference $\Delta f_c$. For the main experiments, we use four seeded uniform-noise references and average the resulting reference-specific explanations. MambaLIFT then propagates multipliers backward through the computational graph using mechanism-aware local rules. Within each VSS block, relevance is decomposed across the selection, recurrent state, and skip pathways of SS2D and further resolved across the scan directions. The resulting relevance is propagated to the input, yielding a spatial attribution map whose sum equals $\Delta f_c$ (reference-relative completeness).

\subsection{Reference-Based Attribution Framework}
\label{sec:objective}

Let $f(x)=[f_1(x),\dots,f_{N_{\mathrm{cls}}}(x)]$ denote the logits of a pretrained VMamba classifier, $c$ the explained class, and $x^0$ a reference. MambaLIFT explains the reference-relative logit difference $\Delta f_c=f_c(x)-f_c(x^0)$. For each variable, $\Delta x=x-x^0$, and relevance is formed from a propagated multiplier as $R_x=m_x\odot\Delta x$. Each local propagation rule satisfies $\sum_i\langle m_{x_i},\Delta x_i\rangle=\langle m_y,\Delta y\rangle$ for an operation $y=\phi(x_1,\ldots,x_n)$.

Operations in the VMamba graph are treated as linear maps, unary nonlinearities, products, additions, or structural tensor operations. For a linear map $y=Wx+b$, the bias cancels in $\Delta y=W\Delta x$ and the multiplier is $m_x=W^\top m_y$. For a unary nonlinearity $y=g(x)$, including Softplus, exponential, SiLU, and GELU, we use the Rescale multiplier
\begin{equation}
m_x=m_y\odot\rho,\qquad
\rho=
\begin{cases}
\Delta y/\Delta x,&\Delta x\neq0,\\
g'(x^0),&\Delta x=0.
\end{cases}
\label{eq:unary}
\end{equation}
When a variable has multiple consumers, their multipliers are summed before further propagation.

\paragraph{Convolutions.}
For convolutional layers, we use the reference-difference form of the generalized $\gamma$-rule of LRP~\cite{montavon2019lrp}, following the formulation given in MambaLRP~\cite{rezaei2024mambalrp}. The contribution of input unit $j$ to output unit $i$ is defined as
\begin{equation}
c_{ij}=w_{ij}\Delta x_j,
\qquad
\Delta x_j=x_j-x_j^0,
\label{eq:conv-contribution}
\end{equation}
where $w_{ij}$ denotes the corresponding convolutional weight and $x_j^0$ is the reference activation. The positive and negative components are
\begin{equation}
[c_{ij}]_{+}=\max(c_{ij},0),
\qquad
[c_{ij}]_{-}=\min(c_{ij},0).
\label{eq:conv-sign-components}
\end{equation}
For each output unit, the propagation branch is selected according to the sign of its reference difference,
\begin{equation}
s_i=
\begin{cases}
+,& \Delta y_i>0,\\
-,& \Delta y_i\leq0,
\end{cases}
\qquad
\Delta y_i=y_i-y_i^0.
\label{eq:conv-sign}
\end{equation}
Biases cancel in the reference difference, giving
\begin{equation}
\Delta y_i=\sum_j c_{ij}.
\label{eq:conv-delta-y}
\end{equation}
We define the effective contribution and its normalization as
\begin{equation}
z_{ij}
=
c_{ij}
+
\gamma[c_{ij}]_{s_i},
\qquad
z_i=\sum_j z_{ij},
\label{eq:gamma}
\end{equation}
where $[c_{ij}]_{s_i}=[c_{ij}]_{+}$ for $s_i=+$ and $[c_{ij}]_{s_i}=[c_{ij}]_{-}$ for $s_i=-$. The propagated relevance is then
\begin{equation}
R_j
=
\sum_i
\frac{z_{ij}}{z_i}R_i.
\label{eq:conv-gamma}
\end{equation}
In multiplier form, let
\begin{equation}
\rho_i=
\begin{cases}
\dfrac{\Delta y_i}{z_i},&z_i\neq0,\\[6pt]
1,&z_i=0.
\end{cases}
\label{eq:conv-rho}
\end{equation}
For $\Delta x_j\neq0$,
\begin{equation}
[w_{ij}\Delta x_j]_{s_i}
=
\Delta x_j
[w_{ij}]_{s_i\operatorname{sgn}(\Delta x_j)},
\end{equation}
where the subscript $s_i\operatorname{sgn}(\Delta x_j)$ is the product of the two signs: the selected part is $[w_{ij}]_{s_i}$ if $\Delta x_j>0$ and $[w_{ij}]_{-s_i}$ if $\Delta x_j<0$. Substituting this identity into Eq.~\eqref{eq:gamma} gives the input multiplier
\begin{equation}
m_{x_j}
=
\sum_i
m_{y_i}\rho_i
\left(
w_{ij}
+
\gamma
[w_{ij}]_{s_i\operatorname{sgn}(\Delta x_j)}
\right).
\label{eq:conv-multiplier}
\end{equation}
Thus, the convolution is propagated as a transposed convolution with sign-dependent effective weights, corresponding to $W+\gamma\max(W,0)$ or $W+\gamma\min(W,0)$ according to the signs of $\Delta x_j$ and $\Delta y_i$. For $\Delta x_j=0$, we have $z_{ij}=0$, so the input receives no contribution from that output and the Linear multiplier $(W^\top m_y)_j$ is retained. We use a single $\gamma$ value for all Conv2d layers, while other linear layers use the Linear rule.

\begin{proposition}[Convolution rule]
\label{prop:gamma}
For $\gamma\geq0$, the multiplier form of Eq.~\eqref{eq:conv-gamma} conserves the reference-relative contribution,
\begin{equation}
\sum_j m_{x_j}\Delta x_j
=
\sum_i m_{y_i}\Delta y_i,
\label{eq:gamma-conservation}
\end{equation}
and reduces to the Linear rule when $\gamma=0$.
\end{proposition}

\begin{pf}
Let
\begin{equation}
S_i=\sum_j[c_{ij}]_{s_i},
\qquad
z_i=\sum_j z_{ij}
=\Delta y_i+\gamma S_i.
\label{eq:proof-denominator}
\end{equation}
Since
\begin{equation}
[c]_+\geq c\geq[c]_-,
\end{equation}
we have
\begin{equation}
\begin{aligned}
s_i=+
&\quad\Longrightarrow\quad
S_i\geq\Delta y_i>0,\\
s_i=-
&\quad\Longrightarrow\quad
S_i\leq\Delta y_i\leq0.
\end{aligned}
\end{equation}
Consequently,
\begin{equation}
\begin{aligned}
s_i=+
&\quad\Longrightarrow\quad
z_i\geq(1+\gamma)\Delta y_i>0,\\
s_i=-
&\quad\Longrightarrow\quad
z_i\leq(1+\gamma)\Delta y_i\leq0.
\end{aligned}
\label{eq:proof-sign}
\end{equation}
Hence $z_i=0$ implies $\Delta y_i=0$, and therefore
\begin{equation}
\rho_i z_i=\Delta y_i
\label{eq:rho-conservation}
\end{equation}
holds for every $i$.

For $\Delta x_j\neq0$, using
\begin{equation}
[w\Delta x]_{\pm}
=
\Delta x\,[w]_{\pm\operatorname{sgn}(\Delta x)},
\end{equation}
we obtain
\begin{equation}
z_{ij}
=
\Delta x_j
\left(
w_{ij}
+
\gamma
[w_{ij}]_{s_i\operatorname{sgn}(\Delta x_j)}
\right).
\label{eq:zij-factorization}
\end{equation}
Thus,
\begin{equation}
m_{x_j}\Delta x_j
=
\sum_i m_{y_i}\rho_i z_{ij}.
\label{eq:multiplier-conservation}
\end{equation}
For $\Delta x_j=0$, both sides vanish because $z_{ij}=0$. Summing over $j$ therefore gives
\begin{equation}
\begin{aligned}
\sum_j m_{x_j}\Delta x_j
&=
\sum_i m_{y_i}\rho_i z_i\\
&=
\sum_i m_{y_i}\Delta y_i,
\end{aligned}
\end{equation}
which establishes local conservation.

For $z_i\neq0$,
\begin{equation}
m_{y_i}\rho_i z_{ij}
=
\frac{z_{ij}}{z_i}R_i,
\qquad
R_i=m_{y_i}\Delta y_i,
\end{equation}
so the multiplier formulation is equivalent to Eq.~\eqref{eq:conv-gamma}.

Finally, when $\gamma=0$,
\begin{equation}
z_{ij}=c_{ij},
\qquad
z_i=\Delta y_i,
\qquad
\rho_i=1,
\end{equation}
which gives
\begin{equation}
m_x=W^\top m_y.
\end{equation}
Hence the rule reduces to the Linear rule.
\end{pf}
\begin{proposition}[Completeness for $\Delta f_c$]
\label{prop:composed-completeness}
If every operation in the VMamba graph satisfies the local conservation identity and the multipliers from all consumers are summed once at shared variables, then the input relevance satisfies
\begin{equation}
\sum_p R_c(p)=\Delta f_c,
\label{eq:objective}
\end{equation}
where $R_c(p)$ is the input relevance summed over colour channels at pixel $p$.
\end{proposition}

\begin{pf}
Starting from the class-logit seed, the graph is processed in reverse topological order. At each operation, the local conservation identity replaces the output contribution by an equal sum of input contributions. Summing consumer multipliers before propagating a shared variable ensures that each pathway is counted exactly once. Thus the total contribution is preserved from the output to the input, yielding $\sum_pR_c(p)=\Delta f_c$.
\end{pf}

Reference-relative completeness is an accounting property and does not by itself imply attribution uniqueness, causal validity, or faithfulness under finite perturbations; these properties are evaluated empirically in Sections~\ref{sec:experiments} and~\ref{sec:exp-ablation}.
\subsection{Reference-Relative Propagation Through Coupled Operations}
\label{sec:layernorm}

LayerNorm couples channels through its shared mean and variance. For $x\in\mathbb{R}^{d}$, let
$\xi=x-\operatorname{mean}(x)$, $v=\operatorname{mean}(\xi^2)$ and $r=(v+\epsilon)^{-1/2}$, where $\epsilon$ is
LayerNorm's numerical-stability constant, so that $\hat x=r\,\xi$, with the scalar $r$ shared by all channels, and $y=\omega\odot\hat x+\beta$. The scale
$\omega$ is a fixed weight and the bias $\beta$ cancels in $\Delta y$. Starting with the downstream multiplier, we
obtain $m_{\hat x}=m_y\odot\omega$. The shared scale receives
\begin{equation}
m_r=\sum_i m_{\hat x,i}\big(\xi^{0}_{i}+\tfrac12\Delta\xi_{i}\big),
\label{eq:ln-mr}
\end{equation}
then $m_v=m_r\,\Delta r/\Delta v$ by Eq.~\eqref{eq:unary} ($r'(v^0)$ when $\Delta v=0$), and the centred input
receives
\begin{align}
m_{\xi}&=\big(r^{0}+\tfrac12\Delta r\big)\,m_{\hat x}
+\tfrac{1}{d}m_v\big(\xi^{0}+\xi\big),
\label{eq:ln-mxc}\\
m_x&=m_{\xi}-\operatorname{mean}(m_{\xi}).
\label{eq:ln-mx}
\end{align}
Eq.~\eqref{eq:ln-mr} and the first term of Eq.~\eqref{eq:ln-mxc} are the two-factor rule~\eqref{eq:2f} for
$\hat x_i=\xi_ir$; the second term passes $m_v\Delta v$ to $\xi$ exactly, since
$\Delta v=d^{-1}\sum_i(\xi^0_i+\xi_i)\Delta\xi_i$; and Eq.~\eqref{eq:ln-mx}, the adjoint of centring, gives
$\langle m_x,\Delta x\rangle=\langle m_\xi,\Delta\xi\rangle$. With $m_v\Delta v=m_r\Delta r$,
\[
\begin{aligned}
\langle m_\xi,\Delta\xi\rangle&=\textstyle\sum_im_{\hat x,i}\big[(r^0+\tfrac12\Delta r)\Delta\xi_i
+(\xi^0_i+\tfrac12\Delta\xi_i)\Delta r\big]\\
&=\langle m_{\hat x},\Delta\hat x\rangle=\langle m_y,\Delta y\rangle,
\end{aligned}
\]
so the local identity holds for any $m_y$. As $\Delta x\to0$, the midpoint factors tend to their values at $x^0$ and
$\Delta r/\Delta v$ to $r'(v^0)$, so $m_x$ tends to the LayerNorm gradient $J^{\top}m_y$.

\paragraph{Two-factor interaction}
For $y=ab$, $\Delta y=b^{0}\Delta a+a^{0}\Delta b+\Delta a\Delta b$. The symmetric rule is
\begin{equation}
R_a=b^{0}\Delta a+\tfrac12\Delta a\Delta b, \qquad
R_b=a^{0}\Delta b+\tfrac12\Delta a\Delta b,
\label{eq:2f}
\end{equation}
with multipliers $m_a=b^{0}+\tfrac12\Delta b$ and $m_b=a^{0}+\tfrac12\Delta a$. It is locally complete
($R_a+R_b=\Delta y$), symmetric, zero-change preserving and equals the two-player Shapley allocation for this product.

\paragraph{Three-factor interaction}
The selective injection $I_t=\tau_tB_tu_t$ is a three-factor product $y=abc$ with
$(a,b,c)=(\tau_t,B_t,u_t)$. Its difference expands as
\begin{equation}
\begin{aligned}
\Delta y ={}& b^0c^0\Delta a+a^0c^0\Delta b+a^0b^0\Delta c\\
&+c^0\Delta a\Delta b+b^0\Delta a\Delta c+a^0\Delta b\Delta c
+\Delta a\Delta b\Delta c.
\end{aligned}
\label{eq:3f-diff}
\end{equation}
The symmetric allocation assigns first-order terms to their factor, half of each pairwise term to its two factors and
one third of the triple term to each:
\begin{equation}
R_a=b^0c^0\Delta a+\tfrac12c^0\Delta a\Delta b+
\tfrac12b^0\Delta a\Delta c+\tfrac13\Delta a\Delta b\Delta c,
\label{eq:3f}
\end{equation}
with the relevances of $b$ and $c$ obtained by permuting the factors. Applied per scan direction, this gives the injection contributions
$R_t^{\tau,\mathrm{inj}},R_t^B,R_t^{u,\mathrm{recurrent}}$.

\begin{proposition}[Three-factor rule]
\label{prop:3f}
Rule~\eqref{eq:3f} is locally complete ($R_a+R_b+R_c=\Delta y$), permutation-symmetric, zero-change preserving and
reduction-consistent (with one factor unchanged, it reduces to Eq.~\eqref{eq:2f} for the other two, scaled by that
factor). For the local three-factor product game of Eq.~\eqref{eq:3f-diff}, it equals the three-player Shapley
allocation and the mean of the three nested decompositions $(ab)c$, $(bc)a$ and $(ca)b$.
\end{proposition}
The Shapley equivalence holds for this local product only; the complete MambaLIFT explanation is not a Shapley
attribution of the network. A single nesting depends on an arbitrary grouping, whereas a sequential allocation assigns interactions to the factor
that changes last; Section~\ref{sec:exp-abl-3f} measures the effect of these alternatives.

\subsection{Reference-Relative Attribution for Selective State-Space Dynamics}
\label{sec:ssm}

The selective recurrence in Eq.~\eqref{eq:recurrence} has a transition pathway $M_t=\bar A_th_{t-1}$ and an injection
pathway $I_t$. In the transition, $A$ is fixed, the exponential uses Eq.~\eqref{eq:unary}, and
$\bar A_th_{t-1}$ uses Eq.~\eqref{eq:2f}. The step size has two consumers, and their multipliers are summed before $\tau_t$ is propagated further:
\begin{equation}
m_{\tau}=m_{\tau,\mathrm{trans}}+m_{\tau,\mathrm{inj}}, \qquad
R^{\tau}_t=m_{\tau}\odot\Delta\tau_t.
\label{eq:tau-merge}
\end{equation}

\paragraph{Temporal propagation}
A contribution entering the state at step $t$ can affect all later states. The two-factor rule gives the backward
state-carrying multiplier
\begin{equation}
m^{h}_{j-1\to j}=\bar A^{0}_j+\tfrac12\Delta\bar A_j, \qquad
m^{h}_{t\to T}=\prod_{j=t+1}^{T}m^{h}_{j-1\to j}.
\label{eq:state-mult}
\end{equation}
State-carrying relevance is kept distinct from transition-parameter relevance and from readout/terminal boundaries.

\paragraph{Readout and skip}
The scan output $y_t=C_th_t+Du_t$, with incoming multiplier $m_{y_t}$, is the sum of the readout $C_th_t$, which uses the symmetric two-factor rule, and the fixed skip $Du_t$,
which uses the Linear rule. Thus $u_t$ also reaches its source through two content pathways,
\begin{equation}
m_u=m_{u,\mathrm{recurrent}}+m_{u,\mathrm{skip}}, \qquad
R^{u}_t=m_u\odot\Delta u_t,
\label{eq:u-merge}
\end{equation}
which are merged within each direction, together with the multipliers of the selection projections
(Section~\ref{sec:ss2d}), before CrossScan's adjoint combines the four directions.

\paragraph{Completeness of the scan.}
For each direction and channel, $h_t$ has two consumers: the readout and the next state transition. The readout contribution to the multiplier of $h_t$ is
\begin{equation}
m^{\mathrm{ro}}_{h_t}
=
\left(
C_t^0+\frac{1}{2}\Delta C_t
\right)m_{y_t},
\label{eq:scan-readout-multiplier}
\end{equation}
while the next transition contributes
\begin{equation}
\left(
\bar A^0_{t+1}
+
\frac{1}{2}\Delta\bar A_{t+1}
\right)
\odot m_{h_{t+1}},
\label{eq:scan-transition-multiplier}
\end{equation}
with the transition contribution set to zero after the last step. Thus,
\begin{equation}
m_{h_t}
=
m^{\mathrm{ro}}_{h_t}
+
\left(
\bar A^0_{t+1}
+
\frac{1}{2}\Delta\bar A_{t+1}
\right)
\odot m_{h_{t+1}}.
\label{eq:scan-multiplier-recurrence}
\end{equation}
Unrolling this recurrence gives
\begin{equation}
m_{h_t}
=
\sum_{T\geq t}
m^h_{t\rightarrow T}
\odot
m^{\mathrm{ro}}_{h_T},
\label{eq:scan-unrolled-multiplier}
\end{equation}
where the transition factors $m^h_{t\rightarrow T}$ are those defined in Eq.~\eqref{eq:state-mult}.

Since the state update decomposes as
\begin{equation}
h_t=M_t+I_t,
\label{eq:scan-state-decomposition}
\end{equation}
the two-factor rule applied to $M_t$ and the three-factor rule applied to $I_t$ give
\begin{equation}
\begin{aligned}
\left\langle
m_{h_t},\Delta h_t
\right\rangle
={}&
\left\langle
\left(
\bar A^0_t+\frac{1}{2}\Delta\bar A_t
\right)
\odot m_{h_t},
\Delta h_{t-1}
\right\rangle
\\
&\quad
+R^{\tau,\mathrm{trans}}_t
+R^{\tau,\mathrm{inj}}_t
+R^B_t
+R^{u,\mathrm{recurrent}}_t .
\end{aligned}
\label{eq:scan-state-completeness}
\end{equation}
Here,
\begin{equation}
R^{\tau,\mathrm{trans}}_t
=
\left\langle
m_{\bar A_t},
\Delta\bar A_t
\right\rangle,
\label{eq:scan-tau-trans}
\end{equation}
where $m_{\bar A_t}=(h^{0}_{t-1}+\tfrac12\Delta h_{t-1})\odot m_{h_t}$ is the two-factor multiplier of $\bar A_t$ in $M_t$; this relevance reaches $\tau_t$ through the exponential transformation in Eq.~\eqref{eq:unary}, with $A$ held fixed, so that $R^{\tau,\mathrm{trans}}_t=m_{\tau,\mathrm{trans}}\Delta\tau_t$. The remaining terms of Eq.~\eqref{eq:scan-state-completeness} are the three-factor allocations of $I_t$ (Eq.~\eqref{eq:3f}) weighted by $m_{h_t}$, with $R^{\tau,\mathrm{inj}}_t=m_{\tau,\mathrm{inj}}\Delta\tau_t$ and $R^{u,\mathrm{recurrent}}_t=m_{u,\mathrm{recurrent}}\Delta u_t$. The readout decomposition is
\begin{equation}
m_{y_t}\Delta y_t
=
\left\langle
m^{\mathrm{ro}}_{h_t},
\Delta h_t
\right\rangle
+
R^C_t
+
R^{u,\mathrm{skip}}_t.
\label{eq:scan-readout-completeness}
\end{equation}
where $R^C_t=\langle(h^{0}_t+\tfrac12\Delta h_t)\,m_{y_t},\Delta C_t\rangle$ is the readout share of $C_t$ and $R^{u,\mathrm{skip}}_t=m_{y_t}D\,\Delta u_t=m_{u,\mathrm{skip}}\Delta u_t$ is the skip contribution.

The first term on the right-hand side of Eq.~\eqref{eq:scan-state-completeness} is precisely the carry contribution to
$\langle m_{h_{t-1}},\Delta h_{t-1}\rangle$. Therefore, summing Eq.~\eqref{eq:scan-state-completeness} over $t$ and using $\Delta h_0=0$ cancels the carry terms, leaving the readout terms $\sum_t\langle m^{\mathrm{ro}}_{h_t},\Delta h_t\rangle$; substituting Eq.~\eqref{eq:scan-readout-completeness} yields
\begin{equation}
\sum_t m_{y_t}\Delta y_t
=
\sum_t
\left(
R^\tau_t
+
R^B_t
+
R^C_t
+
R^u_t
\right),
\label{eq:scan-completeness}
\end{equation}
where $R^\tau_t=R^{\tau,\mathrm{trans}}_t+R^{\tau,\mathrm{inj}}_t$ and $R^u_t=R^{u,\mathrm{recurrent}}_t+R^{u,\mathrm{skip}}_t$ are the merged contributions of Eqs.~\eqref{eq:tau-merge} and~\eqref{eq:u-merge}.

Each pathway therefore enters the total exactly once, as established by Proposition~\ref{prop:composed-completeness}. Omitting or duplicating any pathway changes the total by its corresponding relevance contribution; such an error is detected by the SS2D ledger described in Section~\ref{sec:correctness}.

\subsection{Integration With VMamba's SS2D and VSS Architecture}
\label{sec:ss2d}

SS2D projects the input, applies depthwise convolution and SiLU, forms four row/column scan directions with CrossScan,
runs the selective scan and recombines the directions with CrossMerge, followed by a LayerNorm and an output
projection. The exact adjoints of CrossScan and CrossMerge route multipliers back to their spatial positions. The
scan rules keep the direction-specific contributions separate, and these are summed at their shared source like the
multipliers of any variable with several consumers (Proposition~\ref{prop:composed-completeness}).
The scan input $u_t$ feeds both the content pathway and the selection pathway, in which linear projections of $u_t$
give $\tau_t$ (through a Softplus), $B_t$ and $C_t$.

\paragraph{VSS blocks and configuration}
\label{sec:hier}
\label{sec:config}
VSS block $l$ maps its input $x^{(l)}$ to
\begin{equation}
\begin{aligned}
\tilde x^{(l)}&=x^{(l)}+\mathrm{SS2D}(\mathrm{LN}_1(x^{(l)})),\\
x^{(l+1)}&=\tilde x^{(l)}+\mathrm{MLP}(\mathrm{LN}_2(\tilde x^{(l)})),
\end{aligned}
\label{eq:vss}
\end{equation}
with the residual identity paths retained explicitly. The MLP uses two linear maps with GELU. 
\subsection{Mechanism-Resolved Attribution and Completeness}
\label{sec:mechrep}

The pathways interact in the forward pass, and the local rules allocate these interactions among them. Once the
allocations are fixed, multiplier propagation is linear, so the pathway-specific contributions can be propagated
separately to a block input and summed there without changing their total. Let
$\mathcal M=\mathcal M_{\mathrm{SS2D}}\cup\{\mathrm{identity},\mathrm{MLP}\}$, with
$\mathcal M_{\mathrm{SS2D}}=\{\tau,B,C,u_{\mathrm{recurrent}},u_{\mathrm{skip}}\}$. Here identity is the relevance
carried through both residual identity paths, MLP the relevance that passes the first identity path and then the MLP
branch, and each SS2D source the relevance entering through $\mathrm{LN}_1$ and SS2D, whose output multiplier contains
both downstream paths. 

Scan direction is defined
for SS2D sources, $k\in\mathcal K(\mu)=\{1,\dots,4\}$, and is absent for the identity and MLP branches,
$\mathcal K(\mu)=\{\varnothing\}$. We define $R_c(p,l,k,\mu)$ as the relevance at position $p$ of the input of block
$l$ routed by mechanism $\mu$ in direction $k$; then
\begin{equation}
\sum_{\mu\in\mathcal M}\sum_{k\in\mathcal K(\mu)}R_c(p,l,k,\mu)=R_c^{(l)}(p),\qquad
\sum_pR_c^{(l)}(p)=\Delta f_c.
\label{eq:mechrep}
\end{equation}
For the SS2D branch this gives the exact additive split
$R_{\mathrm{SS2D}}=R_{\mathrm{selection}}+R_{\mathrm{recurrent}}+R_{\mathrm{skip}}$, where selection is
$\{\tau,B,C\}$, recurrent is $u_{\mathrm{recurrent}}$ and skip is $u_{\mathrm{skip}}$, summed over directions.

\begin{figure*}[!t]
\centering
\includegraphics[width=\textwidth]{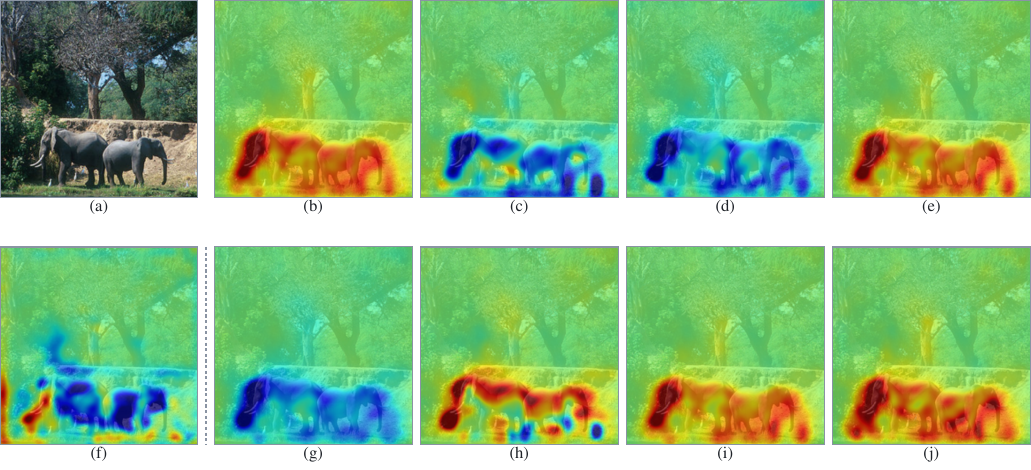}

\caption{Mechanism-aware decomposition of a VMamba-T explanation.
(a) Input image; (b) input relevance; (c)--(e) SS2D relevance assigned to the selection ($\tau$, $B$, $C$),
recurrent-content ($u_{\mathrm{recurrent}}$, the scan input carried through the state) and skip pathways;
(f) state-memory relevance, the part of the relevance of $\tau$, $B$ and $u_{\mathrm{recurrent}}$ read out at least
one scan step after it enters the state; (g)--(j) relevance contributions from the four scan directions.
Maps (c)--(j) are computed for the first VSS block (S0, block~0) on its
$56\times56$ input grid and upsampled to $224\times224$ for display. Each map is scaled separately; red, green,
and blue denote positive, near-zero, and negative relevance, respectively.}

\label{fig:mech-example}
\end{figure*}

For $\tau$, $B$ and $u_{\mathrm{recurrent}}$, the readout at $T=t$ is the local component, whereas $T>t$ gives the
relevance read out at least one scan step after it enters the state through the injection or the transition. This
state-memory (lag) split is a sub-analysis of the same three SS2D sources rather than a fourth additive source, and
introduces no new rule. 
Fig.~\ref{fig:mech-example} illustrates the representation for an image.
The input relevance (b) concentrates on the two elephants and is near zero
on the background. Within the SS2D branch of the first block, the skip pathway
(e) is positive, whereas selection (c) and recurrent content (d) are mainly
negative, so the branch total results from partial cancellation; memory (f)
is also mainly negative on the elephants. By direction, the row-major scan
(g) is negative while the other three scans (h)--(j) are mainly positive,
illustrating distinct directional attribution patterns. As each map is
scaled separately, these observations concern sign and spatial pattern,
not magnitude; Sections~\ref{sec:exp-abl-3f} and~\ref{sec:exp-mech}
quantify these effects.

\paragraph{Reference averaging.}
MambaLIFT is reference-dependent. We use fixed references such as the zero tensor
in normalized input space or an input blur
($\sigma=20$ pixels), and an average over $J$ seeded uniform-noise
references drawn in pixel space and normalized like the input. The explanations
are averaged across references,
$\bar R_c=J^{-1}\sum_jR_c(x;x_j^0)$,
rather than computed using an averaged reference image, and completeness gives
$\sum_p\bar R_c(p)=f_c(x)-J^{-1}\sum_jf_c(x_j^0)$; unless stated otherwise,
$J=4$. With $J=4$, $\bar{R}_c$ is a four-sample Monte Carlo estimate
under the chosen noise-reference distribution. The reference defines the explained quantity and is distinct from the
pixel replacement of the faithfulness evaluation (Section~\ref{sec:exp-setup}).

\section{Experiments}\label{sec:experiments}


\subsection{Experimental Setup}\label{sec:exp-setup}








Table~\ref{tab:exp-setup} summarizes the experimental setup. We evaluate
the released VMamba-T and VMamba-B checkpoints without retraining.
All methods target the pre-softmax logit $f_c$ of the model-predicted
class. GI, SmoothGrad, LRP-LN, and MambaLRP attribute $f_c(x)$,
whereas MambaLIFT and IG attribute the reference-relative difference
$\Delta f_c=f_c(x)-f_c(x^0)$, with $x^0$ denoting the attribution
reference.

\begin{table*}[t]
\centering
\caption{Summary of experimental settings and evaluation protocols.}
\label{tab:exp-setup}
\footnotesize
\renewcommand{\arraystretch}{1.08}
\setlength{\tabcolsep}{5pt}
\begin{tabularx}{\textwidth}{@{}ll>{\raggedright\arraybackslash}X@{}}
\toprule
\textbf{Component} & \textbf{Setting} & \textbf{Parameters / Protocol} \\
\midrule
\textbf{Models} &
VMamba-T, VMamba-B &
\texttt{vssm1\_tiny\_0230s}, \texttt{vssm\_base\_0229};
$d_{\mathrm{state}}=1$, no output gate \\

\textbf{Data} &
ImageNet-1K &
Resize 256, centre crop $224\times224$; 1,000 calibration + 1,000 held-out \\

\textbf{Baselines} &
GI, SmoothGrad, IG, LRP-LN, MambaLRP &
SmoothGrad: 30 samples, $\sigma=0.15$; IG: 30 steps; LRP-LN: LN-rule + LRP-0; MambaLRP: VMamba adaptation, $\gamma=0.25$ \\

\textbf{MambaLIFT} &
Reference-based &
$J=4$ uniform-noise references; $\gamma=0.15$; Gaussian-blur reference ($\sigma=20$ pixels) used in reference-sensitivity analysis \\

\textbf{Faithfulness} &
Pixel flipping &
MoRF/LeRF, 10\% steps; $|\sum_{\mathrm{ch}}R|$ ranking; mean-colour replacement; Gaussian-blur replacement ($\sigma=10$ pixels) \\

\textbf{Statistics} &
Bootstrap + Wilcoxon &
10,000 percentile-bootstrap resamples; paired two-sided Wilcoxon; Holm correction; $\pm0.10$ margin \\

\textbf{Ablations} &
VMamba-T &
1,000 calibration images; frozen configuration; one design choice per run \\
\bottomrule
\end{tabularx}
\end{table*}

\paragraph{Data and evaluation protocol}
We use 2,000 images from the ImageNet-1K validation set, divided into disjoint calibration and held-out sets of 1,000 images each. The sets cover 646 and 617 classes, respectively, with at most five images per class. The calibration set is used for method development, including conservation verification, ablation analyses, reference sensitivity, and selection of attribution settings. The held-out set is not used for these choices and is reserved for the final faithfulness evaluation and statistical comparison. Hyperparameters are selected on VMamba-T before held-out evaluation and then fixed for VMamba-B.

\paragraph{Baselines and references}
We compare MambaLIFT with Gradient$\times$Input (GI), SmoothGrad, Integrated Gradients (IG), LRP-LN, and MambaLRP. MambaLRP uses $\gamma=0.25$, while MambaLIFT uses $\gamma=0.15$ and $J=4$ uniform-noise references. IG is evaluated with either the mean-colour reference or the same noise references used by MambaLIFT. The mean-colour reference corresponds to the zero tensor in normalized input space. For reference-sensitivity analysis, the blurred attribution reference is obtained using a Gaussian filter with $\sigma=20$ pixels; this is distinct from the Gaussian-blur replacement perturbation with $\sigma=10$ pixels used in the secondary faithfulness analysis. Detailed implementation settings will be provided with the accompanying code for reproducibility.

\paragraph{Statistical evaluation}
The attribution reference is distinct from the pixel-replacement strategy used for faithfulness evaluation. The primary comparison uses mean-colour replacement; secondary analyses consider Gaussian-blurred replacement and matched-reference IG. Comparisons are performed separately for VMamba-T and VMamba-B using paired Wilcoxon signed-rank tests~\cite{wilcoxon1945individual}, with Holm correction~\cite{holm1979simple} over the two models within each comparison. Superiority requires adjusted $p<0.05$ and a 95\% bootstrap CI excluding zero; equivalence, assessed with two one-sided tests~\cite{schuirmann1987tost}, requires the 90\% CI to lie within $\pm0.10$, while non-inferiority requires the 95\% CI to lie above $-0.10$. All CIs use 10,000 bootstrap resamples.

\subsection{Implementation Correctness and Conservation}\label{sec:correctness}

We audit MambaLIFT at three levels: local rule consistency, conservation across intermediate model interfaces, and negative-control tests. Each primitive satisfies the local identity, $\sum_i\langle m_{x_i},\Delta x_i\rangle=\langle m_y,\Delta y\rangle$

With the Linear convolution rule ($\gamma=0$), the propagated multiplier converges to the autograd gradient as $\Delta x\rightarrow 0$. For $\gamma>0$, the rule preserves the local identity but not this gradient-limit property.

The released VMamba models use standard float32 computation. For auditing, we evaluate the same parameters and inputs using a float64 computation tape, in which the SS2D selective scan is computed step by step and every hidden state is stored, together with float64 multiplier propagation. Conservation and interface consistency are assessed in this audit computation; agreement with the original float32 computation checks that the audit reproduces the model's forward computation. Validation proceeds from individual primitives through the recurrent step, scan, SS2D, VSS blocks, stem, downsampling layers, and classifier. This covers 138 audited interfaces for VMamba-T and 194 for VMamba-B, together with an 11-interface ledger within each SS2D. Across calibration images, the median relative difference between the reference-relative logit difference $\Delta f_c$ computed by the audit and by the original float32 model is $1.2\times10^{-7}$ for VMamba-T and $1.8\times10^{-7}$ for VMamba-B.

We additionally test the gradient limit using \(\gamma=0\) and \(x^0=x-\varepsilon v\), where \(v\) is a random direction, and introduce four implementation faults: swapped CrossScan directions, omission of the recurrent-content and skip pathways in the \(u\) branch, an incorrect SiLU derivative at \(\Delta x=0\), and a float32 selective-scan backward pass. All four faults are detected. Notably, the incorrect SiLU derivative preserves conservation because it affects only the zero-increment case, but violates the gradient-limit test. Thus, conservation alone is insufficient to establish implementation correctness, consistent with the broader finding that visual plausibility alone does not establish model dependence in saliency methods~\cite{adebayo2018sanity}.

Gradient$\times$Input (GI) and MambaLRP attribute the class logit \(f_c(x)\), whereas MambaLIFT attributes the reference-relative difference \(\Delta f_c=f_c(x)-f_c(x^0)\). Accordingly, the conservation results in Table~\ref{tab:audit-conservation} should be interpreted as agreement with each method's respective attribution target, rather than as a direct comparison of conservation errors across methods. This distinction is important when interpreting conservation discrepancies. In LRP, relevance can be absorbed by additive constants, yielding
\[
f_c(x)-\sum_i R_i=\sum_\ell D_\ell,
\]
where \(D_\ell\) denotes the relevance absorbed by layer-specific constants, including LayerNorm shifts and convolution, MLP, and classifier biases~\cite{bach2015lrp,montavon2019lrp}. Such constants cancel when computing MambaLIFT's reference-relative difference.

The original MambaLRP study evaluates Vim-S, which uses RMSNorm without an additive shift and contains no MLP blocks, reducing the number of such constant terms. Accordingly, the conservation gap observed on VMamba should be interpreted in the context of architectural differences, propagation conventions, and the quantity being explained. Table~\ref{tab:audit-conservation} summarizes the audit results.

\begin{table*}[t]
\centering
\caption{Implementation audit and conservation comparison. Internal conservation of the float64 MambaLIFT computation across audited interfaces; agreement between the MambaLIFT relevance sum and the target $\Delta f_c$ computed by the original float32 VMamba computation; and agreement between baseline relevance sums and their stated output-logit targets.}
\label{tab:audit-conservation}
\footnotesize
\begin{tabular}{@{}p{0.24\textwidth}p{0.35\textwidth}p{0.35\textwidth}@{}}
\toprule
Evaluation & VMamba-T & VMamba-B\\
\midrule
Audited interfaces & 138 & 194\\
Worst relative audit error & $1.7\times10^{-15}$ & $1.5\times10^{-15}$\\
\midrule
MambaLIFT, target $\Delta f_c$ & rel.\ $1.2\times10^{-7}$; $R^2_{y=x}=1.000$ & rel.\ $1.8\times10^{-7}$; $R^2_{y=x}=1.000$\\
\midrule
GI, target $f_c(x)$ & rel.\ $0.82$; $R^2_{y=x}=-91.8$ & rel.\ $0.86$; $R^2_{y=x}=-326.7$\\
MambaLRP, target $f_c(x)$ & rel.\ $0.31$; $R^2_{y=x}=-32.8$ & rel.\ $0.19$; $R^2_{y=x}=-38.6$\\
\bottomrule
\end{tabular}
\end{table*}

In Table~\ref{tab:audit-conservation}, the relevance sum \(\sum m\odot\Delta\) at each interface is compared with the corresponding mathematically defined target using absolute error and scale-normalized relative error:
\[
\mathrm{Absolute\ error}
=
\left|\sum_i m_i\odot\Delta x_i-\mathrm{target}\right|,
\]
\[
\mathrm{Relative\ error}
=
\frac{\left|\sum_i m_i\odot\Delta x_i-\mathrm{target}\right|}
{\max\left(\sum_i|m_i\odot\Delta x_i|,|\mathrm{target}|\right)}.
\]

For the calibration-set conservation experiment (Table~\ref{tab:audit-conservation}), MambaLIFT uses the main experimental configuration, \(\gamma=0.15\) with \(J=4\) noise references. For each method, the relevance sum is compared with its explained quantity \(y\): \(f_c(x)\) for GI and MambaLRP, and \(f_c(x)-J^{-1}\sum_jf_c(x_j^0)\) for the averaged MambaLIFT explanation. These targets are computed using the original float32 model. The per-image relative error, \(\left|\sum_i R_i-y\right|/|y|\), is summarized by its median. Agreement with the identity line is quantified across the \(n\) images using
\[
R^2_{y=x}
=1-\frac{\sum_n\left(\sum_iR_{n,i}-y_n\right)^2}
{\sum_n(y_n-\bar y)^2}.
\]
It can be negative when the squared discrepancies between relevance sums and their targets exceed the variance of the target values across images.

\subsection{Qualitative Explanations}\label{sec:exp-qual}

\begin{figure*}[!t]
\centering
\includegraphics[width=\textwidth]{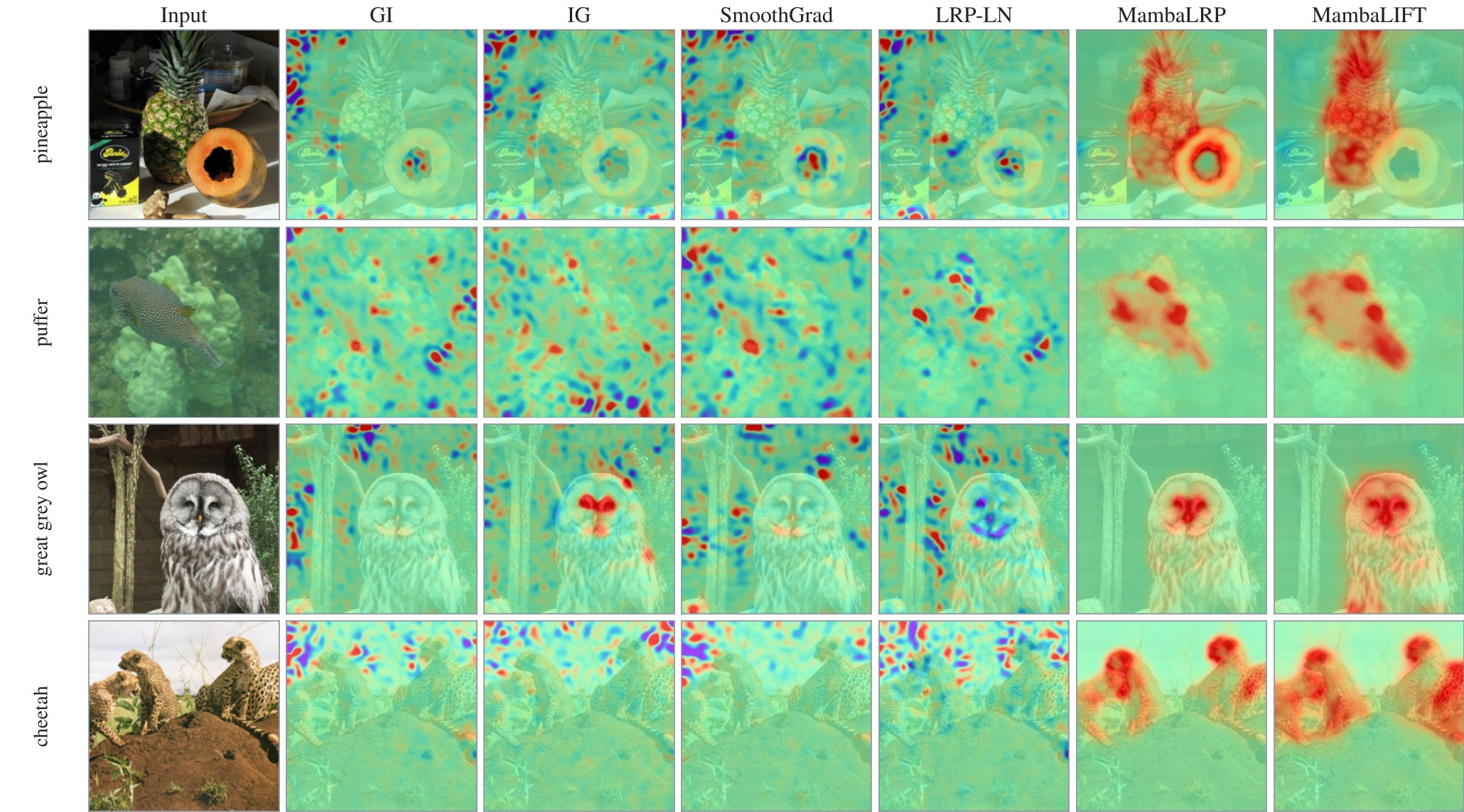}
\caption{Qualitative attribution maps of VMamba-T for four images from the held-out evaluation set. Columns show the input image and explanations from Gradient$\times$Input (GI), Integrated Gradients (IG), SmoothGrad, LRP-LN, MambaLRP, and MambaLIFT.}
\label{fig:exp-qual}
\end{figure*}

Fig.~\ref{fig:exp-qual} compares explanations generated by different methods for four illustrative images from the held-out evaluation set, using VMamba-T and the model-predicted class as the attribution target. To facilitate visual comparison of the spatial patterns, channel-wise relevance is summed to obtain a two-dimensional spatial map and a Gaussian filter with \(\sigma=4\) pixels is applied uniformly to all methods to suppress fine-scale pixel variation while retaining the larger spatial structures of the explanations. Because attribution methods can produce relevance maps with different absolute scales, each map is independently normalized by its 99th percentile of absolute relevance; this preserves the within-map spatial and sign structure while preventing a small number of extreme values from determining the displayed intensity range. Relevance polarity is retained throughout visualization, with positive, near-zero, and negative values displayed in red, green, and blue, respectively, to distinguish supporting, negligible, and opposing contributions. These transformations are used solely to improve the comparability and readability of the qualitative visualizations and are not used in any quantitative evaluation; all faithfulness and statistical analyses are computed from the corresponding raw relevance maps.

In these examples, MambaLIFT generally produces more spatially concentrated and target-focused attributions than MambaLRP. For the pineapple image (1st row), both methods highlight the fruit, but MambaLIFT produces a more coherent attribution pattern with less relevance extending into the surrounding scene, while MambaLRP shows stronger relevance outside the primary fruit region. For the puffer-fish image (2nd row), both methods localize the fish, but MambaLIFT produces a more concentrated response around the target. The owl example (3rd row) shows comparable localization between the two methods. For the cheetah image (4th row), MambaLIFT concentrates relevance on the animals and their body regions, whereas MambaLRP produces more spread-out foreground attribution. Overall, MambaLIFT shows a clearer tendency toward spatially concentrated, target-focused explanations in these examples. These observations complement the quantitative faithfulness results.

\subsection{Faithfulness Evaluation}\label{sec:exp-faith}

We evaluate attribution faithfulness using pixel-flipping experiments on the 1,000 images of the held-out evaluation set. For each image, pixels are ranked by the absolute channel-summed attribution magnitude, \( \left|\sum_{\mathrm{ch}} R\right| \). Under the most-relevant-first (MoRF) ordering, pixels with the largest attribution magnitudes are replaced first; under the least-relevant-first (LeRF) ordering, replacement proceeds in the reverse order. Let \(A_F^{\mathrm{MoRF}}\) and \(A_F^{\mathrm{LeRF}}\) denote the areas under the corresponding model-response curves. We quantify the separation between the two orderings using
\[
\Delta A_F = A_F^{\mathrm{LeRF}}-A_F^{\mathrm{MoRF}}.
\]
A larger \(\Delta A_F\) indicates that the attribution ranking better distinguishes pixels whose replacement strongly affects the model response from those whose replacement has a smaller effect. Within each comparison, all methods use the same pixel-replacement strategy for a consistent evaluation protocol.

\subsubsection{Held-Out Faithfulness Results}

The configurations selected during calibration were fixed before evaluation on the held-out set. Mean-colour replacement was used as the prespecified primary pixel-flipping protocol for all methods (Table~\ref{tab:exp-faith}). All quantitative faithfulness results reported in this subsection are computed exclusively on the held-out evaluation set.

As reported in Table~\ref{tab:exp-faith}, MambaLIFT achieves the highest mean \(\Delta A_F\) among the evaluated methods for both VMamba configurations and outperforms MambaLRP in the prespecified primary comparison. The improvement is larger for VMamba-T than for VMamba-B, but remains positive for both configurations. Thus, under mean-colour replacement on the held-out evaluation set, MambaLIFT achieves higher \(\Delta A_F\) than MambaLRP for both evaluated configurations. The robustness analysis below repeats the comparison using blurred replacement.

\begin{table*}[t]
\centering
\caption{Faithfulness on the 1,000-image held-out evaluation set under the primary mean-colour replacement protocol. Higher $\Delta A_F$ is better; 95\% bootstrap CIs are shown.}
\label{tab:exp-faith}
\footnotesize
\setlength{\tabcolsep}{12pt}
\begin{tabular*}{\textwidth}{@{\extracolsep{\fill}}lrcrrrr@{}}
\toprule
Method & \multicolumn{1}{c}{$\Delta A_F$} & 95\% CI & Median & $\Delta A_F>0$ (\%) & $ A_F^{\mathrm{MoRF}}$ & $ A_F^{\mathrm{LeRF}}$\\
\midrule
\multicolumn{7}{@{}l}{\textit{VMamba-T}}\\
Random                 & $0.0101$ & $[-0.0131,\,0.0337]$ & $-0.0002$ & 49.9 & $5.4676$ & $5.4777$\\
GI                     & $0.1173$ & $[0.0062,\,0.2294]$ & $0.0436$ & 51.1 & $5.3976$ & $5.5148$\\
SmoothGrad             & $1.3588$ & $[1.2580,\,1.4567]$ & $1.1938$ & 80.1 & $4.8092$ & $6.1680$\\
IG, mean-colour baseline        & $0.4195$ & $[0.3064,\,0.5318]$ & $0.2702$ & 58.0 & $5.2187$ & $5.6382$\\
IG, noise baseline ($J=4$)      & $2.5393$ & $[2.4420,\,2.6342]$ & $2.3274$ & 97.2 & $4.1247$ & $6.6640$\\
LRP-LN                 & $0.9441$ & $[0.8266,\,1.0586]$ & $0.7633$ & 69.3 & $5.0108$ & $5.9550$\\
MambaLRP               & $3.8756$ & $[3.7765,\,3.9712]$ & $3.8953$ & 99.7 & $3.4992$ & $7.3748$\\
MambaLIFT (ours)       & $\mathbf{4.1599}$ & $[4.0612,\,4.2555]$ & $\mathbf{4.2248}$ & \textbf{99.8} & $\mathbf{3.3656}$ & $\mathbf{7.5256}$\\
\midrule
\multicolumn{7}{@{}l}{\textit{VMamba-B}}\\
Random                 & $0.0133$ & $[-0.0097,\,0.0365]$ & $0.0058$ & 50.8 & $6.0206$ & $6.0339$\\
GI                     & $-0.0604$ & $[-0.1680,\,0.0484]$ & $-0.1838$ & 46.4 & $5.8948$ & $5.8343$\\
SmoothGrad             & $1.0678$ & $[0.9666,\,1.1677]$ & $0.8727$ & 74.7 & $5.3945$ & $6.4624$\\
IG, mean-colour baseline        & $0.4877$ & $[0.3787,\,0.5963]$ & $0.3584$ & 58.3 & $5.6334$ & $6.1211$\\
IG, noise baseline ($J=4$)      & $2.6842$ & $[2.5863,\,2.7819]$ & $2.5478$ & 98.4 & $4.4467$ & $7.1309$\\
LRP-LN                 & $0.8776$ & $[0.7650,\,0.9901]$ & $0.6929$ & 68.6 & $5.4686$ & $6.3462$\\
MambaLRP               & $3.7155$ & $[3.6159,\,3.8148]$ & $3.7000$ & 99.7 & $4.0007$ & $\mathbf{7.7161}$\\
MambaLIFT (ours)       & $\mathbf{3.8102}$ & $[3.7084,\,3.9120]$ & $\mathbf{3.7878}$ & \textbf{99.8} & $\mathbf{3.8504}$ & $7.6607$\\
\bottomrule
\end{tabular*}
\end{table*}

\subsubsection{Replacement-Robustness Analysis}

Because pixel-flipping results depend on the replacement strategy, we repeat the held-out evaluation using Gaussian-blurred replacement (\(\sigma=10\) pixels), keeping the attribution maps and reference configurations unchanged. As summarized in Table~\ref{tab:exp-faith-summary}, MambaLIFT remains competitive with MambaLRP, but its superiority under mean-colour replacement does not generalize uniformly across both VMamba configurations. VMamba-T meets the prespecified non-inferiority criterion, whereas VMamba-B meets the prespecified equivalence criterion. MambaLIFT also achieves higher $\Delta A_F$ than IG with the matched noise baseline under both replacements.

\begin{table*}[t]
\centering

\caption{Paired differences in $\Delta A_F$ between MambaLIFT and the comparison methods on the held-out set. Positive values favour MambaLIFT; confidence intervals are percentile-bootstrap intervals. $p$-values are Holm-adjusted Wilcoxon signed-rank tests.}

\label{tab:exp-faith-summary}
\footnotesize
\setlength{\tabcolsep}{3pt}
\begin{tabular}{@{}llrccrlrcr@{}}
\toprule
 & & \multicolumn{5}{c}{vs.\ MambaLRP} & \multicolumn{3}{c}{vs.\ IG, noise baseline ($J=4$)}\\
\cmidrule(lr){3-7}\cmidrule(l){8-10}
Replacement & Model & Difference & 95\% CI & 90\% CI & $p$ & Outcome & Difference & 95\% CI & $p$\\
\midrule
Mean colour       & VMamba-T & $+0.2844$ & $[0.2225,\,0.3467]$  & $[0.2312,\,0.3359]$  & $4.5\times10^{-17}$ & Superior     & $+1.6206$ & $[1.5412,\,1.7014]$ & $<10^{-100}$\\
Mean colour       & VMamba-B & $+0.0948$ & $[0.0318,\,0.1577]$  & $[0.0418,\,0.1480]$  & $0.0044$            & Superior     & $+1.1261$ & $[1.0519,\,1.1987]$ & $<10^{-100}$\\
\midrule
Blur, $\sigma=10$ & VMamba-T & $+0.0605$ & $[0.0025,\,0.1178]$  & $[0.0116,\,0.1085]$  & $0.051$             & Non-inferior & $+1.2554$ & $[1.1763,\,1.3357]$ & $<10^{-100}$\\
Blur, $\sigma=10$ & VMamba-B & $+0.0296$ & $[-0.0292,\,0.0874]$ & $[-0.0190,\,0.0781]$ & $0.425$             & Equivalent   & $+1.0207$ & $[0.9530,\,1.0874]$ & $<10^{-100}$\\
\bottomrule
\end{tabular}

\smallskip
\begin{minipage}{\linewidth}\raggedright\scriptsize
\noindent\textit{Note: Differences are computed from the unrounded evaluation scores; component scores are displayed to four decimal places, so subtracting the displayed values may differ by $0.0001$.}
\end{minipage}
\end{table*}

\subsection{Computational Cost}\label{sec:exp-cost}

We measure computational runtime on an NVIDIA GeForce RTX 5060 GPU with 8\,GB of memory, using three repetitions per image after warm-up. Table~\ref{tab:exp-runtime} reports the median wall-clock time per image and the peak GPU memory required. Under the reported implementation, MambaLIFT incurs higher computational cost than the comparison methods, primarily due to float64 execution and separate backward propagation for each reference. MambaLIFT uses float64 arithmetic for multiplier propagation, whereas the released VMamba forward computation uses float32 arithmetic; the reported runtime therefore reflects the implementations and numerical precisions used in the experiments rather than a matched-precision comparison.

\begin{table}[!t]
\centering
\caption{Computational cost per explanation for VMamba-T and VMamba-B.}
\label{tab:exp-runtime}

\small
\setlength{\tabcolsep}{2pt}
\renewcommand{\arraystretch}{1.05}

\begin{tabular}{lrrrr}
\toprule
& \multicolumn{2}{c}{VMamba-T} & \multicolumn{2}{c}{VMamba-B}\\
\cmidrule(lr){2-3}\cmidrule(lr){4-5}
Method & s/image & GB & s/image & GB\\
\midrule
Forward pass
& 0.012 & 0.02
& 0.018 & 0.05 \\

GI
& 0.027 & 0.15
& 0.040 & 0.35 \\

SmoothGrad (30)
& 0.205 & 1.21
& 0.522 & 3.22 \\

IG (30)
& 0.191 & 1.18
& 0.510 & 3.20 \\

LRP-LN
& 0.037 & 0.12
& 0.055 & 0.28 \\

MambaLRP
& 0.056 & 0.20
& 0.082 & 0.48 \\

\midrule
MambaLIFT, one reference
& 2.17 & 0.54
& 3.23 & 1.20 \\

MambaLIFT, $J=4$, sequential
& 8.66 & 0.54
& 12.9 & 1.20 \\

MambaLIFT, $J=4$, batch of 4
& 3.17 & 2.26
& 6.77 & 5.14 \\

\bottomrule
\end{tabular}
\end{table}

\section{Ablation Study}\label{sec:exp-ablation}

The ablation studies examine propagation rules, reference choices, temporal propagation, and pathway composition using VMamba-T. Unless stated otherwise, experiments use the frozen configuration (\(\gamma=0.15\), four noise references) on the 1,000-image calibration set, with each experiment changing one propagation rule or design choice while keeping all other settings and the forward pass fixed. The unmodified implementation is first checked for reproducibility and conservation. Conservation and attribution-map agreement are evaluated separately, since different propagation rules can conserve the same explained quantity while producing different attribution maps.

\subsection{Interaction Allocation and Pathway Composition}\label{sec:exp-abl-3f}

For the selective-scan interaction \(\tau_t B_t u_t\), we compare the direct three-factor allocation with nested two-factor, sequential, and first-order alternatives. As reported in Table~\ref{tab:exp-3f}, all exact variants conserve the explained quantity but produce different attribution maps. The nested decompositions closely match the direct rule (\(\rho_S=0.9948\)–\(0.9968\)), whereas sequential allocations yield larger deviations. Averaging the three nested decompositions reproduces the direct rule to within \(2.9\times10^{-16}\), while the first-order approximation fails to capture the full interaction. These results show that completeness alone does not uniquely determine the attribution map. The direct symmetric allocation corresponds to the Shapley-value allocation of the three-factor interaction.

\begin{table}[!t]
\centering

\caption{Three-factor attribution ablation for the selective-scan injection $\tau_tB_tu_t$.}

\label{tab:exp-3f}
\footnotesize
\setlength{\tabcolsep}{4pt}
\begin{tabular}{@{}lrrrr@{}}
\toprule
Allocation & $\sum R/\Delta f_c$ & $\rho_S$ & $\mathrm{IoU}_{10}$ & rel.\ $\ell_2$\\
\midrule
Direct                  & 1         & 1      & 1      & 0\\
\midrule
\addlinespace
Nested $(\tau B)u$            & 1           & 0.9955 & 0.9256 & 0.05927\\
Nested $\tau(Bu)$             & 1           & 0.9968 & 0.9361 & 0.04946\\
Nested $(\tau u)B$            & 1           & 0.9948 & 0.9226 & 0.06161\\
\midrule
Mean of nested$^{a}$          & 1           & 1.0000 & 1.0000 & $2.9{\cdot}10^{-16}$\\
\midrule
\addlinespace
Seq.\ $\tau\to B\to u$        & 1           & 0.8254 & 0.6076 & 0.4336\\
Seq.\ $u\to B\to\tau$         & 1           & 0.8491 & 0.6310 & 0.3742\\
\midrule
\addlinespace
First order only              & $-0.007165$ & 0.1512 & 0.2824 & 1.046\\
\bottomrule
\end{tabular}

\smallskip
\begin{minipage}{\linewidth}\raggedright\scriptsize
$^{a}$Evaluated on one image as an identity check.
\end{minipage}
\end{table}

At shared sources, MambaLIFT propagates each incoming pathway separately and merges the resulting multipliers at the shared node. Across the 14 VMamba-T blocks, the single-node additivity identity holds to within \(7\times10^{-16}\). However, in Table~\ref{tab:exp-merge}, no individual pathway recovers the complete attribution. Signed accounting ratios \(R/\Delta f_c\) range from 0.030 to 2.443, with Spearman rank correlations (\(\rho_S\)) ranging from 0.435 to 0.935. Because these are signed contributions from incomplete pathways, the accounting ratio is not constrained to the interval \([0,1]\). The affected pathways include step-size transition and injection, content and selection, CrossMerge directions, and the identity, SS2D, and MLP residual branches. These results show that local conservation does not make incoming pathways interchangeable; their contributions must be composed jointly at shared sources.

\begin{table}[!t]
\centering

\caption{Pathway-composition ablations across the 14 VMamba-T blocks.}

\label{tab:exp-merge}
\small
\setlength{\tabcolsep}{2pt}

\begin{tabular}{lrrrr}
\toprule
Ablation & Retained pathway & $\sum R/\Delta f_c$ & $\rho_S$ & $\mathrm{IoU}_{10}$ \\
\midrule
Complete & All pathways & 1.000 & 1.000 & 1.000 \\
\midrule
Step size $\tau$ & Injection $\tau Bu$ & 2.443 & 0.935 & 0.722 \\
                 & Transition $\exp(\tau A)$ & 0.352 & 0.763 & 0.513 \\
\midrule
Content $u$ & Recurrent injection & 0.264 & 0.716 & 0.517 \\
            & $D$-skip & 0.510 & 0.753 & 0.534 \\
\midrule
Scan fan-out & Content $u$ & 0.916 & 0.904 & 0.656 \\
             & Selection $(\tau,B,C)$ & 0.139 & 0.707 & 0.490 \\
\midrule
CrossMerge & $k=1$ & 0.249 & 0.673 & 0.427 \\
           & $k=2$ & 0.030 & 0.435 & 0.315 \\
           & $k=3$ & 0.617 & 0.770 & 0.524 \\
           & $k=4$ & 0.387 & 0.772 & 0.506 \\
\midrule
Residual & Identity + MLP & 0.159 & 0.679 & 0.427 \\
         & Identity + SS2D & 0.049 & 0.614 & 0.426 \\
\bottomrule
\end{tabular}
\end{table}

\subsection{Temporal Propagation}\label{sec:exp-abl-temporal}

MambaLIFT propagates the state multiplier backward using the midpoint coefficient \(\bar A_t^0+\tfrac12\Delta\bar A_t\) (Eq.~\eqref{eq:state-mult}). Alternative one-sided rules based on \(\bar A_t\) or \(\bar A_t^0\) also conserve the explained quantity but yield different attribution maps, with Spearman correlations of \(\rho_S=0.9371\) and \(0.5420\), and relative \(\ell_2\) distances of \(0.2313\) and \(1.259\), respectively. We also examine truncated-memory variants, in which a source entering the recurrent state at step \(t\) receives relevance only from readouts up to \(t+\omega\). Because truncation omits later contributions, these variants do not conserve the full logit difference; accordingly, \(\sum R/\Delta f_c\) is a signed accounting ratio, not a bounded relevance fraction. In Table~\ref{tab:exp-temporal}, increasing \(\omega\) brings the truncated attribution closer to the midpoint result. At \(\omega=256\), the correlation reaches \(\rho_S=0.9979\) and \(\mathrm{IoU}_{10}=0.9869\), with an accounting ratio of \(1.009\). These findings demonstrate that temporal attributions depend on both the state-transition propagation rule and the extent of recurrent memory included.

\begin{table}[!t]
\centering

\caption{Temporal-propagation ablations for carry coefficients and recurrent memory windows.}
\label{tab:exp-temporal}
\footnotesize
\setlength{\tabcolsep}{4pt}
\begin{tabular}{@{}lrrrr@{}}
\toprule
Variant & $\sum R/\Delta f_c$ & $\rho_S$ & $\mathrm{IoU}_{10}$ & rel.\ $\ell_2$\\
\midrule
Midpoint           & 1      & 1      & 1      & 0\\
\addlinespace
Carry $\bar A_t$         & 1      & 0.9371 & 0.7354 & 0.2313\\
Carry $\bar A^{0}_t$     & 1      & 0.5420 & 0.4153 & 1.259\\
\addlinespace
Window $\omega=0$             & 0.7132 & 0.7819 & 0.5424 & 0.5402\\
Window $\omega=1$             & 0.6810 & 0.8101 & 0.5853 & 0.5143\\
Window $\omega=4$             & 0.6156 & 0.8634 & 0.6551 & 0.4942\\
Window $\omega=16$            & 0.7222 & 0.9337 & 0.7804 & 0.3367\\
Window $\omega=64$            & 1.014  & 0.9841 & 0.9281 & 0.07884\\
Window $\omega=256$           & 1.009  & 0.9979 & 0.9869 & 0.02034\\
\bottomrule
\end{tabular}
\end{table}

\subsection{Reference Sensitivity}\label{sec:exp-reference}

MambaLIFT attributes the reference-relative logit difference, \(\Delta f_c=f_c(x)-f_c(x^0)\), and is therefore sensitive to the choice of attribution reference. We assess this sensitivity on the 1,000-image calibration set using the Linear convolution rule (\(\gamma=0\)) to isolate the effect of the reference from the convolution-rule allocation. Each reference configuration is compared with the mean-colour reference using Spearman rank correlation and \(\mathrm{IoU}_{10}\). In Table~\ref{tab:abl-reference}, all configurations satisfy completeness, while yielding substantially different attribution maps.

Averaging four noise references improves agreement with the mean-colour baseline compared with using a single noise reference. Accordingly, the four-noise-reference configuration, together with \(\gamma=0.15\), is adopted for the main faithfulness evaluation. This sensitivity to reference choice is consistent with prior studies of attribution baselines and explanation reliability~\cite{sturmfels2020baselines,kindermans2019unreliability}.

\begin{table}[!t]
\centering
\caption{Reference sensitivity of MambaLIFT across eight reference configurations. Spearman correlation $\rho_S$ and IoU$_{10}$ are measured relative to the mean-colour reference.}

\label{tab:abl-reference}
\small
\setlength{\tabcolsep}{4pt}
\begin{tabular}{lrrr}
\toprule
Reference & $\Delta f_c$ & $\rho_S$ & $\mathrm{IoU}_{10}$ \\
\midrule
Mean colour & 8.944 & 1.0000 & 1.0000 \\
Black & 8.928 & 0.1085 & 0.2819 \\
White & 8.983 & 0.2157 & 0.3240 \\
Blur, $\sigma=5$ & 5.283 & 0.1111 & 0.1257 \\
Blur, $\sigma=10$ & 7.756 & 0.1940 & 0.1634 \\
Blur, $\sigma=20$ & 8.529 & 0.2977 & 0.2121 \\
Uniform noise & 8.945 & 0.2636 & 0.2034 \\
Mean of $J=4$ noise references & 8.945 & 0.3860 & 0.2512 \\
\bottomrule
\end{tabular}
\end{table}

\subsection{Mechanism-Level Accounting and Intervention Analysis}\label{sec:exp-mech}

\begin{figure*}[!t]
\centering

\begin{minipage}[t]{0.485\textwidth}
    \centering
    \includegraphics[width=\linewidth]{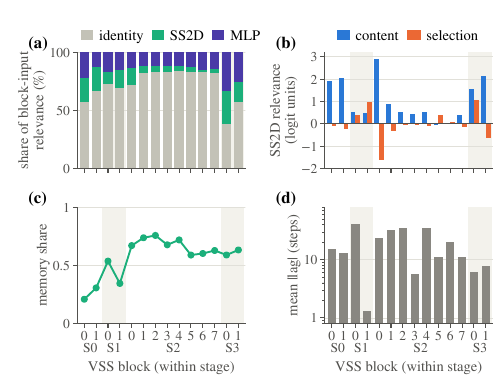}

    \vspace{2pt}
    \refstepcounter{figure}
    \label{fig:exp-depthprofile}
    \parbox{\linewidth}{%
        \textbf{Figure \thefigure:}
    
Block-wise mechanism profiles of VMamba-T under the four-noise-reference configuration (\(J=4\), \(\gamma=0.15\)). (a) Share of block-input relevance assigned to the identity, SS2D, and MLP pathways. (b) SS2D relevance assigned to the content and selection pathways, shown in logit units. (c) Fraction of scan relevance attributed to state-memory propagation. (d) Relevance-weighted mean absolute readout lag, measured in scan steps. The horizontal axis indexes VSS blocks within stages S0–S3.
}
    
\end{minipage}
\hfill
\begin{minipage}[t]{0.485\textwidth}
    \centering
    \includegraphics[width=\linewidth]{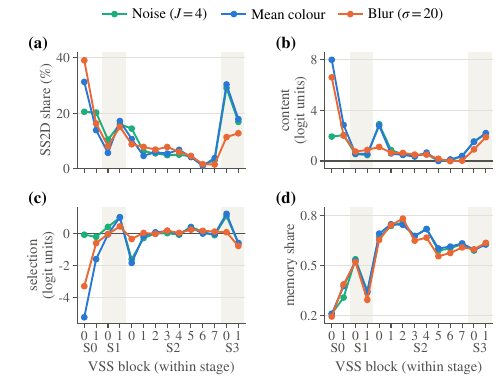}

    \vspace{2pt}
    \refstepcounter{figure}
    \label{fig:exp-mechref}
    \parbox{\linewidth}{%
        \textbf{Figure \thefigure:}
       
        Block-wise mechanism profiles of VMamba-T for three reference configurations: four noise references (\(J=4\)), mean-colour references, and Gaussian-blur references (\(\sigma=20\) pixels). (a) SS2D share of block-input relevance. (b) Content-pathway relevance in logit units. (c) Selection-pathway relevance in logit units. (d) Fraction of scan relevance attributed to state-memory propagation. The horizontal axis indexes VSS blocks within stages S0–S3.
}
    
\end{minipage}

\end{figure*}

We analyze how the reference-relative logit difference is distributed across the VMamba-T hierarchy and how mechanism-level relevance relates to interventions that replace individual mechanisms with their reference values. Unless stated otherwise, the analysis uses VMamba-T with \(\gamma=0.15\) on the 1,000-image calibration set; the global memory intervention uses the first 250 images. Fig.~\ref{fig:exp-depthprofile} decomposes the same \(\Delta f_c\) into identity, SS2D, and MLP contributions, with a mean of 8.94 under the four-noise-reference configuration (\(J=4\)). Fig.~\ref{fig:exp-mechref} compares mechanism-level profiles across the four-noise-reference, mean-colour, and Gaussian-blur (\(\sigma=20\)) configurations.

\subsubsection{Mechanism Profiles}

The identity path accounts for the largest share of block-input relevance in every block (38--84\%). Within SS2D, content relevance is positive across all 14 blocks, whereas selection relevance is negative in nine. The memory share increases from 0.21 in the first block to 0.59--0.76 across S2, remaining approximately 0.6 in S3. In several S1 and S2 blocks, relevance-weighted temporal lags extend to 24--41 scan steps. Across reference configurations, SS2D, content, and memory exhibit consistent block-wise rankings (\(\rho_S=0.88\)--\(0.99\)), while selection is less stable (\(\rho_S=0.61\)--\(0.80\)), particularly in early blocks.

\subsubsection{Intervention Analysis}

For mechanism $\mu$ in VSS block $l$, let $f_c(x;\mu_l\leftarrow\mu_l^0)$ denote the class logit obtained by replacing $\mu$ with its reference value while leaving all other computations unchanged. The intervention effect is $E_{l,\mu}=f_c(x)-f_c(x;\mu_l\leftarrow\mu_l^0)$, which is compared with the corresponding MambaLIFT relevance $R_{l,\mu}$. Selection is intervened on jointly through $\tau_t$, $B_t$, and $C_t$, and on each variable individually. Content and skip interventions revert $u_t$ in $\tau_tB_tu_t$ and $Du_t$, respectively, while the full-SS2D intervention reverts all four variables.

For memory, the scan is recomputed using $h_t=\bar A_t h^0_{t-1}+\tau_tB_tu_t$, preventing differences from being carried beyond their point of entry. The resulting effect is compared with the state-memory relevance from Section~\ref{sec:mechrep}. The global memory effect $E_{\mathrm{glob}}$ applies the same intervention across all scans and blocks and is compared with the memory share $1-\sum_pR^{\omega=0}_c(p)/\Delta f_c$ defined in Section~\ref{sec:exp-abl-temporal}. For the noise configuration, $E$ and $R$ are averaged over the $J=4$ references.

As reported in Table~\ref{tab:abl-interventions}, relevance--intervention correlations are modest for content and full-SS2D, weaker for skip, and negative for memory and selection. In the gradient limit ($\gamma=0$, $x^0=x-\varepsilon v$), relevance and intervention effects agree to first order, with a maximum relative difference of $1.4\times10^{-3}$.
\begin{table}[!t]
\centering
\caption{Mechanism-level intervention results for VMamba-T ($\gamma=0.15$). Local entries are Spearman correlations between mechanism relevance and logit drop; global entries report the memory share of $\Delta f_c$ and the normalized intervention effect $E_{\mathrm{glob}}/\Delta f_c$.}
\label{tab:abl-interventions}
\small
\setlength{\tabcolsep}{5pt}
\begin{tabular}{lrrr}
\toprule
Mechanism & Noise & Mean colour & Blur \\
\midrule
\multicolumn{4}{l}{\textit{Local: Spearman $\rho_S$ of $R$ and $E$}}\\
Memory & $-0.11$ & $-0.10$ & $-0.03$ \\
Selection $(\tau,B,C)$ & $-0.09$ & $-0.12$ & $-0.01$ \\
$\tau$ alone & 0.25 & 0.19 & 0.02 \\
$B$ alone & 0.13 & 0.15 & 0.30 \\
$C$ alone & $-0.16$ & $-0.22$ & $-0.04$ \\
Content injection $u$ & 0.29 & 0.33 & 0.39 \\
Content skip $Du_t$ & 0.06 & 0.18 & 0.09 \\
Whole SS2D branch & 0.24 & 0.33 & 0.26 \\
\midrule
\multicolumn{4}{l}{\textit{Global memory}}\\
Memory share of $\Delta f_c$ & 0.30 & 0.10 & 0.56 \\
$E_{\mathrm{glob}}/\Delta f_c$ & $-0.29$ & $-0.25$ & 0.15 \\
Intervention-relevance $\rho_S$ & 0.44 & 0.42 & 0.45 \\
\bottomrule
\end{tabular}
\end{table}

Reverting the whole SS2D branch differs from the sum of its content, skip, and selection interventions by 0.54--0.93 logit units. For the first 250 images, memory accounts for 30\%, 10\%, and 56\% of $\Delta f_c$ under noise, mean-colour, and blur references, respectively, while the normalized global memory effect $E_{\mathrm{glob}}/\Delta f_c$ is $-0.29$, $-0.25$, and $0.15$. Intervention--relevance correlations are only 0.42--0.45.

\subsection{Selection of
\texorpdfstring{$\gamma$}{gamma}
and Attribution Reference}
\label{sec:hyperparameter-selection}

The tunable parameters—the convolution-rule parameter \(\gamma\) and attribution reference for MambaLIFT, and the LRP-\(\gamma\) parameter for MambaLRP—were selected on the VMamba-T calibration set and fixed before held-out evaluation. The search grids and selection protocol were prespecified. The configuration with the highest mean \(\Delta A_F\) under the prespecified mean-colour pixel-flipping protocol was selected. Because all attribution configurations were selected exclusively on the calibration set, the held-out estimates do not participate in hyperparameter selection. As shown in Table~\ref{tab:calib}, MambaLIFT selects four noise references with \(\gamma=0.15\), while MambaLRP selects \(\gamma=0.25\). LRP-LN, GI, SmoothGrad, and the random-map baseline have no parameters tuned on the calibration set.

\begin{table}[!t]
\centering
\caption{Calibration results on the 1,000-image VMamba-T calibration set under the prespecified mean-colour pixel-flipping protocol. Values are mean $\Delta A_F$ for the predefined attribution configurations. MambaLIFT is evaluated across convolution-rule $\gamma$ values and reference configurations; MambaLRP is evaluated across its LRP-$\gamma$ grid.}

\label{tab:calib}
\footnotesize
\setlength{\tabcolsep}{2.6pt}
\begin{tabular}{@{}lrrrr@{}}
\toprule
 & \multicolumn{3}{c}{MambaLIFT, reference} & \\
\cmidrule(lr){2-4}
$\gamma$ & Mean colour & Blur, $\sigma=20$ & Noise, $J=4$ & MambaLRP\\
\midrule
0    & 2.0453 & 2.7639 & 2.7888 & 1.8912\\
0.15 & 3.7525 & 3.7447 & 4.0976 & 3.7907\\
0.25 & 3.8153 & 3.7059 & 4.0677 & 3.8475\\
0.30 & 3.8310 & 3.6642 & 4.0667 & 3.8362\\
0.35 & 3.8089 & 3.6316 & 4.0690 & 3.8181\\
\midrule
IG, baseline & 0.3191 & 2.1122 & 2.5208 & \\
\bottomrule
\end{tabular}
\end{table}

\section{Conclusion}\label{sec:conclusion}

We presented MambaLIFT, a reference-based attribution framework for VMamba that decomposes reference-relative class-logit differences across spatial positions, network blocks, scan directions, and computational pathways. Its multiplier-based rules preserve the explained logit difference to floating-point precision across all audited interfaces in VMamba-T and VMamba-B. On the held-out evaluation set, MambaLIFT significantly outperforms MambaLRP in mean \(\Delta A_F\) under the prespecified mean-colour pixel-flipping protocol for both configurations. Under Gaussian-blurred replacement, however, the results support non-inferiority for VMamba-T and equivalence for VMamba-B, indicating that the advantage depends on the replacement strategy.

Conservation does not uniquely determine attribution maps: interaction allocation, temporal propagation, pathway composition, and reference choice all affect the explanations. Moreover, conserved relevance does not necessarily predict finite output changes under mechanism-level interventions. The results identify multiplicative interactions, recurrent memory, scan directions, and shared pathways as important components of the attribution analysis. Generalization to larger state dimensions, gated SS2D variants, and other selective state-space architectures remains to be established. Future work will extend MambaLIFT to these settings and improve the efficiency of backward-scan propagation.

\appendices
\counterwithin{table}{section}
\counterwithin{figure}{section}




\balance
\bibliographystyle{IEEEtran}
\bibliography{mambalift-refs}

\end{document}